# Mammogram Edge Detection Using Hybrid Soft Computing Methods

I. Laurence Aroquiaraj, K. Thangavel

**Abstract**— Image segmentation is a crucial step in a wide range of method image processing systems. It is useful in visualization of the different objects present in the image. In spite of the several methods available in the literature, image segmentation still a challenging problem in most of image processing applications. The challenge comes from the fuzziness of image objects and the overlapping of the different regions. Detection of edges in an image is a very important step towards understanding image features. There are large numbers of edge detection operators available, each designed to be sensitive to certain types of edges. The Quality of edge detection can be measured from several criteria objectively. Some criteria are proposed in terms of mathematical measurement, some of them are based on application and implementation requirements. Since edges often occur at image locations representing object boundaries, edge detection is extensively used in image segmentation when images are divided into areas corresponding to different objects. This can be used specifically for enhancing the tumor area in mammographic images. Different methods are available for edge detection like Roberts, Sobel, Prewitt, Canny, Log edge operators. In this paper a novel algorithms for edge detection has been proposed for mammographic images. Breast boundary, pectoral region and tumor location can be seen clearly by using this method. For comparison purpose Roberts, Sobel, Prewitt, Canny, Log edge operators are used and their results are displayed. Experimental results demonstrate the effectiveness of the proposed approach.

————————— ◆ —————————

## 1 INTRODUCTION

Breast cancer has been one of the major causes of death among women since the last decades and it has become an emergency for the healthcare systems of industrialized countries. This disease became a commonest cancer among women. If the cancer can be detected early, the options of treatment and the chances of total recovery will increase. Intra-operative diagnosis of the disease has steadily become more important with respect to the recent introduction of sentinel lymph node biopsy. Image segmentation is referred to as the procedure in which the input image is divided into meaningful regions in such a way that the output image will consist of a set of labeled region describing the input image.

The potency of digital mammography for detecting of breast cancer is currently under Investigation. In medical science, mammography image is to be a cornerstone for examining breast cancer in human. Even if image output obtained in scanning process with X ray is frequent blueprint and unclearness for edge of image. Screen-film mammography has limited detection ability for low contrast lesions in dense breasts. This limitation poses a problem for the estimated 40% of women with dense breast who undergo mammography [1].

Thus, to be required a process for improving the contrast of mammography images was a digital image processing. The researchers have been done several methods for improving the capability of mammography image to detect the abnormality. Enhancement of contrast of the mammograms has been done by the researchers in the past.

Edge detection is a critical element in image processing, since edges contain a major function of image information. The function of edge detection is to identify the boundaries of homogeneous regions in an image based on properties such as intensity and texture. Many edge detection algorithms have been developed based on computation of the intensity gradient vector, which, in general, is sensitive to noise in the image.

The rest of this paper is organized as follows. Section (2) reviews edge detection operators. Section (3) classifies the edge detection algorithms. Section (4) discusses proposed edge detection algorithm. Section, (5) the experimental result of many edge detections obtained and finally in section (7) the discussion and conclusion are given.

## 2 Review of Previous Work

In the past two decades several algorithms were developed to extract the contour of homogeneous regions within digital mammogram image. A lot of the attention is focused to edge detection, being a crucial part in most of the algorithms.

- Laurence Aroquiaraj was IEEE Member, Assistant Professor, Department of Computer Science, Periyar University, Salem – 636 011, Tamil Nadu, India(corresponding author to provide phone: +918807058505; e-mail: laurence.raj@gmail.com).

- Prof. Dr. K. Thangavel, was IEEE Member, Professor and Head, Department of Computer Science, Periyar University, Salem – 636 011, Tamil Nadu, India (e-mail: drktvelu@yahoo.com).

Classically, the first stage of edge detection (e.g. the gradient operator, Robert operator, the Sobel operator, the Prewitt operator) is the evaluation of derivatives of the image intensity. Smoothing filter and surface fitting are used as regularization techniques to make differentiation more immune to noise. Raman Maini and J. S. Sobel [2] evaluated the performance of the Prewitt edge detector for noisy image and demonstrated that the Prewitt edge detector works quite well for digital image corrupted with Poisson noise whereas its performance decreases sharply for other kind of noise.

Ferrari et al. (2004) [3] has proposed a new method using Gabor wavelets for the identification of the pectoral muscle in medio-lateral oblique (MLO) mammograms based upon a multiresolution technique. The magnitude value of each pixel was propagated in the direction of the phase after computing the magnitude and phase images using a vector-summation procedure. The resulting image was then used to detect the relevant edges and true pectoral muscle edge. However, in many cases, cancer is not easily detected by the eyes.

Bellotti et al. [4] characterized ROI by means of textural features computed from the gray level co-occurrence matrix (GLCM), also known as spatial gray level dependence (SGLD) matrix. Varela et al. [5] used features based on the iris filter output, together with gray level, texture, contour-related and morphological features. Yuan et al. [6] used three groups of features in their study. The first group included features characterizing spiculation, margin, shape and contrast of the lesion. Sahiner et al. [7] developed an algorithm for extracting spiculation feature and circumsribed margin feature. Both features had high accuracy for characterizing mass margins according to BI-RADS descriptors.

Timp and Karssemeijer[8] proposed temporal feature set consisted of complete set of single view features together with temporal features. Timp et al. [9] designed two kinds of temporal features: difference features and similarity features. Difference features measured changes in feature values between corresponding regions in the prior and the current view. Similarity features measured whether two regions are comparable in appearance. Fauci et al. [10] extracted 12 features from segmented masses. Some features gave the geometrical information, others provided shape parameters. The criterion for feature selection was based on morphological differences between pathological and healthy regions. Rangayyan et al. [11] proposed methods to obtain shape features from the turning angle functions of contours. Features are useful in the analysis of contours of breast masses and tumors because of their ability to capture diagnostically important details of shape related to spicules and lobulations.

Davis, L. S. [12] has suggested Gaussian pre convolution for this purpose. However, all the Gaussian and Gaussian-like smoothing filters, while smoothing out the noise, also remove genuine high frequency edge features, degrade localization and degrade the detection of low-contrast edges.

Zhao Yu-qian et al. [13] proposed a novel mathematic morphological algorithm to detect lungs CT medical image edge. They showed that this algorithm is more efficient for medical image denoising and edge detecting than the usually used template-based edge detection algorithms such as Laplacian of Gaussian operator and Sobel edge detector, and general morphological edge detection algorithm such as morphological gradient operation and dilation residue edge detector. Fesharaki, M.N. and Hellestrand, G.R [14] presented a new edge detection algorithm based on a statistical approach using the student t-test. They selected a 5x5 window and partitioned into eight different orientations in order to detect edges.

However there is no such loss in the fuzzy based method described here. Research has clearly demonstrated that methods involving Gaussian filtering suffer from problems such an edge displacement, vanishing edges and false edges [15]. Another problem faced by few methods like the anisotropic diffusion lies in obtaining the locations of semantically meaningful edges at coarse scales generated by convoluting images with Gaussian kernels [16]. Methods that involve simple scan line approach are not able to detect all the edges due to limitation of the methodology to trace only the horizontal and vertical neighbours [17] of a point. Fuzzy logic is a powerful problem-solving methodology with a myriad of applications in embedded control and information processing [18]. Fuzzy provides a remarkably simple way to draw definite conclusions from vague, ambiguous or imprecise information. In a sense, fuzzy logic resembles human decision making with its ability to work from approximate data and find precise solutions.

The fuzzy relative pixel value algorithm has been developed with the knowledge of vision analysis with low or no illumination [19], thus making this method optimized for application requiring such methods. The method helps us to detect edges in an image in all cases due to subjection of pixel values to an algorithm involving host of fuzzy conditions for edges associated with an image. The purpose of this paper is to present a new methodology for image edge detection which is undoubtedly one of the most important operations related to low level computer vision, in particular within area of feature extraction with plethora of techniques, each based on a new methodology, having been published. The method described here uses a fuzzy based logic model with the help of which high performance is achieved along with simplicity in resulting model [20]. Fuzzy logic helps to deal with problems with imprecise and vague information and thus helps to create a model for image edge detection as presented here [21] displaying the accuracy of fuzzy methods in digital image processing [22].

# 3 Edge Detection Techniques
## 3.1 Robert Edge Detector

The calculation of the gradient magnitude of an image is obtained by the partial derivatives $G_x$ and $G_y$ at every pixel location. The simplest way to implement the first order partial derivative is by using the Roberts cross gradient operator. Therefore

$$G_x = f(i, j) - f(i+1, j+1) \tag{1}$$
$$G_y = f(i+1, j) - f(i, j+1) \tag{2}$$

## 3.2 Prewitt Edge detector

The Prewitt edge detector is a much better operator than Roberts's operator. This operator having a 3 x 3 masks deals better with the effect of noise. An approach using the masks of size 3 x 3 is given below, the arrangement of pixels about the pixels[i, j]. The partial derivatives of the Prewitt operator are calculated as

$$G_x = (a6 + ca5 + a4) - (a0 + ca1 + a2) \tag{3}$$
$$G_y = (a2 + ca3 + a4) - (a0 + ca7 + a6) \tag{4}$$

The constant c implies the emphasis given to pixels closer to the centre of the mask. $G_x$ and $G_y$ are the approximation at [i, j]. Setting c=1, the Prewitt operator is obtained. Therefore the Prewitt masks are as follows these masks have longer support. They differentiate in one direction and average in the other direction, so the edge detector is less vulnerable to noise.

## 3.3 Sobel Edge Detector

The Sobel edge detector is very much similar to the Prewitt edge detector. The difference between the both is that the weight of the centre coefficient is 2 in the Sobel operator. The partial derivatives of the Sobel operator are calculated as

$$G_x = (a6 + 2a5 + a4) - (a0 + 2a1 + a2) \tag{5}$$
$$G_y = (a2 + 2a3 + a4) - (a0 + 2a7 + a6) \tag{6}$$

Although the Prewitt masks are easier to implement than Sobel masks, the later has better noise suppression characteristics.

## 3.4 Laplacian of Gradient (LoG) Operator Edge Detector

The Laplacian of an image $f(x, y)$ is a second order derivative defined as:

$$\nabla^2 f = \frac{\partial^2 f}{\partial x^2} + \frac{\partial^2 f}{\partial y^2} \tag{7}$$

The Laplacian of Gradient (LoG) is usually used to establish whether a pixel is on the dark or light side of an edge.

## 3.5 Canny Edge Detector

Canny technique is very important method to find edges by isolating noise from the image before find edges of image, without affecting the features of the edges in the image and then applying the tendency to find the edges and the critical value for threshold. The algorithmic steps for canny edge detection technique are follows:

Algorithm: Canny Edge Detection
Step1: Convolve image $f(r,c)$ with a Gaussian function to get smooth image $f(r,c)$
$$f(r,c) = f(r,c) * G(r,c,6) \tag{8}$$
Step 2: Apply first difference gradient operator to compute edge strength then edge magnitude and direction are obtain as before.
Step 3: Apply non-maximal or critical suppression to the gradient magnitude.
Step 4: Apply threshold to the non-maximal suppression image.

# 4 The Proposed Edge Detection algorithms
## 4.1 Fuzzy Canny Edge Detector

The hybrid of Fuzzy and Canny edge detection technique is very important method to find edges by isolating noise from the image before find edges of image, without affecting the features of the edges in the image and then applying the tendency to find the edges and the critical value for threshold. The algorithmic steps for canny edge detection technique are Fuzzy provides a remarkably simple way to draw definite conclusions from vague, ambiguous or imprecise information. In a sense, fuzzy logic resembles human decision making with its ability to work from approximate data and find precise solutions.

Algorithm: Fuzzy Canny Edge Detection
Step 1: Convolve image $f(r,c)$ with a fuzzy logic to get fuzzified image $f(r,c)$
Step 2: Fuzzified image $f(r,c)$ with a Gaussian function to get smooth image $f(r,c)$
Step 3: Apply first difference gradient operator to compute edge strength then edge magnitude and direction are obtain as before.
Step 4: Apply non-maximal or critical suppression to the gradient magnitude.
Step 5: Apply threshold to the non-maximal suppression image.

## 4.2 Fuzzy Relative Pixel Edge Detector

The Algorithm begins with reading an MxN image. The first set of nine pixels of a 3x3 window is chosen with central pixel having values (2, 2). After the initialization, the pixel values are subjected to the fuzzy conditions for edge existence shown in Fig.2.(a-i). After the subjection of the pixel values to the fuzzy conditions the algorithm generates an intermediate image. It is checked whether all pixels have been checked or now, if not then first the horizontal coordinate pixels are checked. If all horizontal pixels have been checked the vertical pixels are checked else the horizontal pixel is incremented to retrieve the next set of pixels of a window. In this manner the window shifts and checks all the pixels in one horizontal line then increments to check the next vertical location.

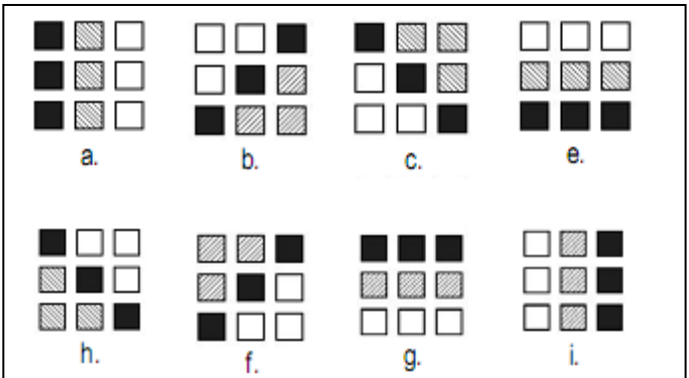

Fig 1(a-i). Fuzzy conditions have been displayed

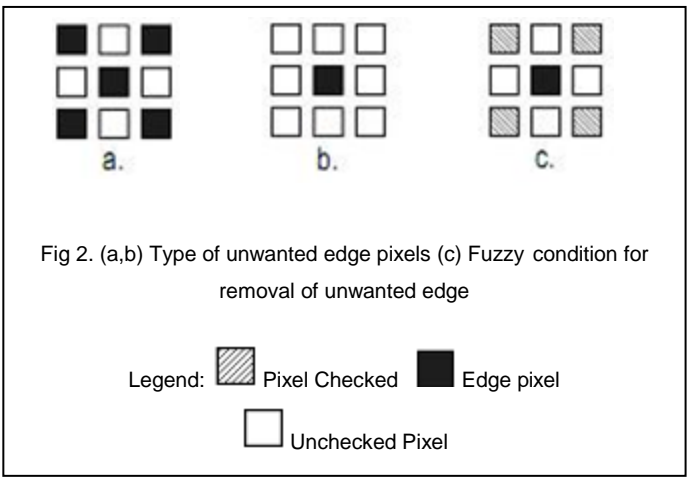

Fig 2. (a,b) Type of unwanted edge pixels (c) Fuzzy condition for removal of unwanted edge

Legend: ▨ Pixel Checked  ■ Edge pixel  ☐ Unchecked Pixel

After edge highlighting image is subjected to another set of condition with the help of which the unwanted parts of the output image of type shown in Fig.2.(a-b) are removed to generate an image which contains only the edges associated with the input image. Let us now consider the case of the fuzzy condition displayed in Fig.1. (g). for an input image A and an output image B of size MxN pixels respectively we have the following set of conditions that are implemented to detect the edges pixel values.

### 4.3 Fuzzy Edge Detection Based on Pixel's Gradient and Standard Deviation Values (SDGD Edge Detector)

The gradient and standard deviation of pixels value, edges are separately extracted and then based on fuzzy logic, final decision about whether each pixel is edge or not is made. Problematic results could be gained if each of the methods be used solely. It may causes on identifying of edge pixels as non-edge pixels and vice versa.

Algorithm: Fuzzy based Gradient and Standard Deviation Value Edge Detection

Step 1: Convolve image $f(r,c)$ with a fuzzy logic to get fuzzified image $f(r,c)$

Step 2: Fuzzified image $f(r,c)$ with a gradient values are computed by the Sobel operator $G_x$ and $G_y$ function to get smooth image $f(r,c)$.

Step 2: Apply the smooth image $f(r,c)$ gray level standard deviation to compute over adjacent neighbourhood pixels.

Step 3: Similarly pixels with standard deviation values is greater than a threshold value are edge.

Step 4: Apply threshold to the non-maximal suppression image using 3x3 window size.

## 5 Results and Discussion

Obtaining real mammogram images (322 images) for carrying out research is highly difficult due to privacy issues, legal issues and technical hurdles. Hence the Mammography Image Analysis Society (**MIAS**) database (ftp://peipa.essex.ac.uk) is used in this paper to study the efficiency of the proposed image segmentation and evaluated using mammography images. The proposed system was tested with different mammogram images, its performance being compared to that of the other edge detection operators.

The objective methods used to measure the performance of edge detectors using signal to noise ratio and mean square error between the edge detectors images and the original one. The objective methods borrowed from digital signal processing and information theory, and provide us with equations that can be used to measure the amount of error in a processed image by comparison to known image. Although the objective methods are widely used, are not necessarily correlated with our perception of image quality. For instance, an image with a low error as determined by an objective measure may actually look much worse than an image with a high error metric. Table 1 explains the objective measures are the root-mean-square error, $e_{RMS}$, the root-meansquare signal-to-noise ratio, $SNR_{RMS}$, and the peak signal-to-noise ratio, $SNR_{PEAK}$ and Contrast Improved Index (CII).

TABLE 1
OBJECTIVE METHOD MEASURES

| Statistical Measurement | Formula |
|---|---|
| MSE | $\dfrac{\sum (f(i,j) - F(i,j))^2}{MN}$ |
| $e_{RMS}$ | $\sqrt{\dfrac{\sum (f(i,j) - F(i,j))^2}{MN}}$ |
| $SNR_{RMS}$ | $10 \log_{10} \dfrac{\sigma^2}{\sigma_e^2}$ |
| $SNR_{PEAK}$ | $20 \log_{10} \dfrac{255}{RMSE}$ |
| CII (Contrast Improved Index) | $\dfrac{I_{max} - I_{min}}{I_{max} + I_{min}}$ |

If the value of $e_{RMS}$ is low and the values of $SNR_{RMS}$ and $SNR_{PEAK}$ are larger than the enhancement approach is better. Fig. 3 Different Edge Detection Methods and Proposed Fuzzy based Edge Detection Methods using mdb002, mdb067, mdb171, mdb240 and mdb320 mammogram images.

TABLE 2
PERFORMANCE RATES OF EDGE DETECTORS FOR 322 MAMMOGRAM IMAGES

| | $e_{RMS}$ | $SNR_{RMS}$ | $SNR_{AVERAGE}$ | $SNR_{PEAK}$ | CII |
|---|---|---|---|---|---|
| Sobel | 87.33 | 101.65 | -0.03 | 8.07 | 1.00 |
| Canny | 87.28 | 91.9 | -0.03 | 8.08 | 1.00 |
| Prewitt | 87.33 | 101.71 | -0.03 | 8.07 | 1.00 |
| Robert | 87.38 | 102.93 | -0.03 | 8.07 | 1.00 |
| Log | 87.34 | 95.04 | -0.03 | 8.07 | 1.00 |
| Fuzzy | 86.96 | 88.88 | -0.03 | 8.11 | 1.00 |
| Fuzzy canny | 87.34 | 92.52 | -0.03 | 8.07 | 1.00 |
| Fuzzy Relative Pixel | 87.35 | 106.63 | -0.03 | 8.07 | -0.63 |
| Sdgd | 87.34 | 107.62 | -0.03 | 8.07 | -0.01 |

It is observed from the Table 2 that according to edge detectors provide the performance rates for overall 322 mammogram images are equal except values of MSE, $e_{RMS}$, $SNR_{RMS}$ and $SNR_{PEAK}$. The proposed Fuzzy based Edge detectors provides the best performance rate. Fig. 4 shows the performance analysis of different edge detectors.

the best percentage of total number of white pixels. Fig. 5 shows the performance analysis of different edge detectors.

TABLE 3
PERCENTAGE OF WHITE PIXELS VALUES FOR EDGE DETECTORS

| | Mdb002 | | Mdb067 | | Mdb171 | | Mdb240 | | Mdb320 | |
|---|---|---|---|---|---|---|---|---|---|---|
| | Total White Pixels | % | Total White Pixels | % | Total White Pixels | % | Total White Pixels | % | Total White Pixels | % |
| Sobel | 280 | 1.76 | 269 | 1.69 | 238 | 1.50 | 268 | 1.69 | 204 | 1.28 |
| Canny | 639 | 4.02 | 852 | 5.37 | 596 | 3.75 | 701 | 4.42 | 629 | 3.96 |
| Prewitt | 280 | 1.76 | 266 | 1.68 | 235 | 1.48 | 263 | 1.66 | 203 | 1.28 |
| Robert | 371 | 2.34 | 248 | 1.56 | 263 | 1.66 | 214 | 1.35 | 210 | 1.32 |
| Log | 501 | 3.16 | 519 | 3.27 | 389 | 2.45 | 427 | 2.69 | 359 | 2.26 |
| Fuzzy | 3325 | 20.94 | 5981 | 37.67 | 5145 | 32.41 | 6295 | 39.65 | 6897 | 43.44 |
| Fuzzy canny | 647 | 4.07 | 867 | 5.46 | 784 | 4.94 | 782 | 4.93 | 1021 | 6.43 |
| Fuzzy Relative Pixel | 5137 | 32.36 | 7033 | 44.30 | 6587 | 41.49 | 7988 | 50.31 | 8532 | 53.74 |
| Sdgd | 5645 | 35.56 | 7486 | 47.15 | 7196 | 45.33 | 8658 | 54.54 | 9237 | 58.18 |

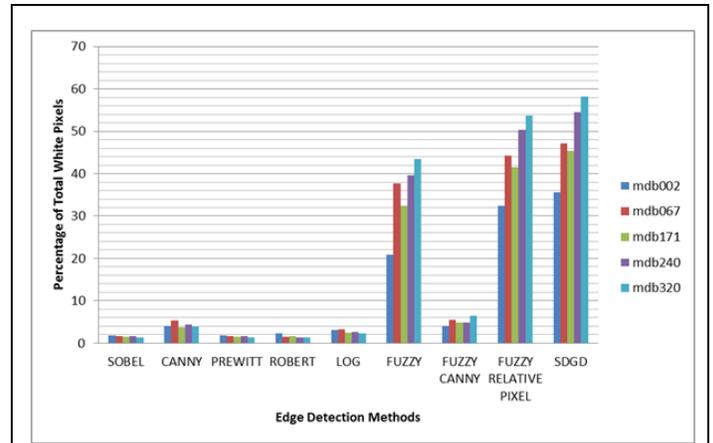

Fig. 5 Performance analysis of different Edge Detectors using Percentage of White Pixels

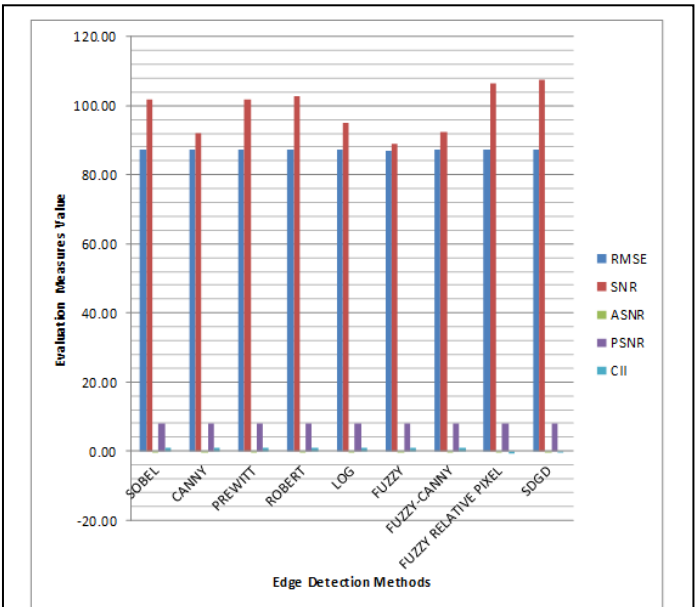

Fig. 4 Performance analysis of different Edge Detectors

It is observed from the Table 3 that according to edge detectors provide the percentage of total number of white pixels in selected mammogram images for different edge detectors. The proposed Fuzzy based Edge detectors provides

## 6 CONCLUSION

In this paper, better algorithm has been proposed to improve the detection of edges by using fuzzy rules. This algorithm is adaptable to various environments. The weights associated with each fuzzy rule were tuned to allow good results to be obtained while extracting edges of the image, where contrast varies a lot from one region to another. During the performance tests, however, all parameters were kept constant. Experimental results show the higher quality and superiority of the extracted edges compared to the other methods in the literature such as Sobel, Robert, and Prewitt. To achieve good result, some parameters and thresholds are experimentally set. Improving fuzzy system performance by the ways such as using different kind of input, rough fuzzy hybridization techniques also need to be investigated in future works.

## ACKNOWLEDGEMENT

The first author of this paper acknowledges UGC-Minor Research Project (No.F.41-1361/2012(SR)) for the financial support.

| Edge DetectionMethods | Mdb002 | Mdb067 | Mdb171 | Mdb240 | Mdb320 |
|---|---|---|---|---|---|
| Canny | 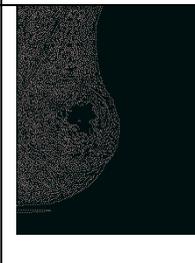 | 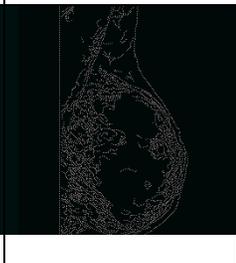 | 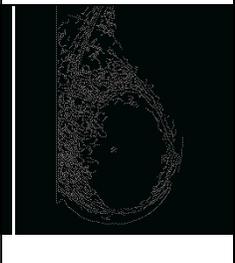 | 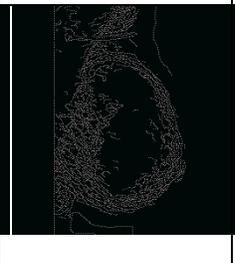 | 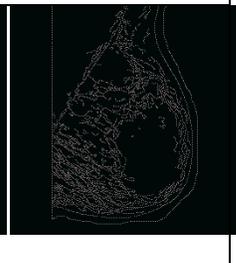 |
| Fuzzy | 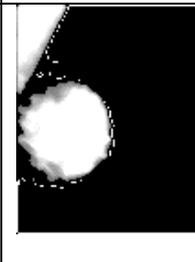 | 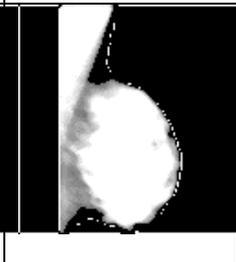 | 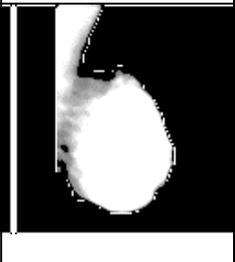 | 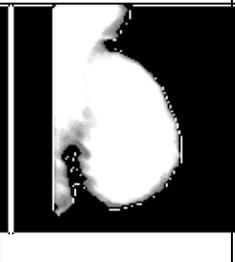 | 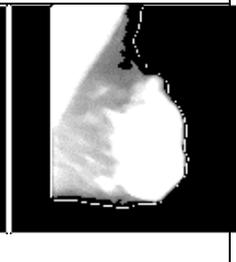 |
| Fuzzy Canny | 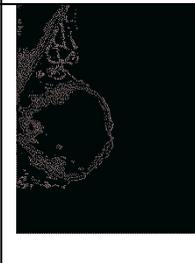 | 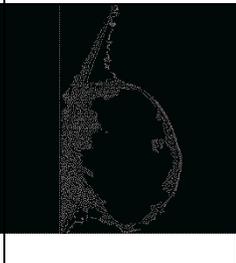 | 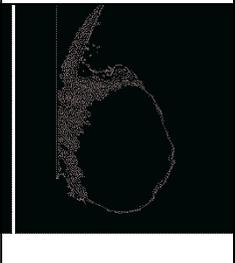 | 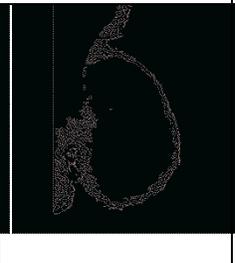 | 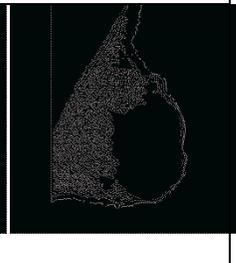 |
| Log | 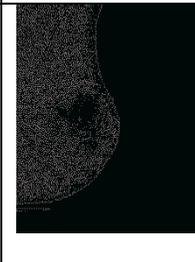 | 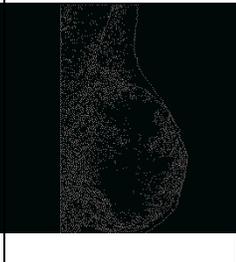 | 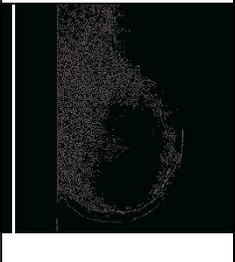 | 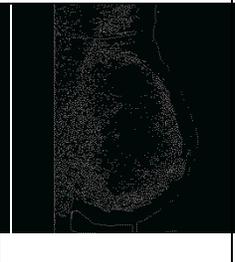 | 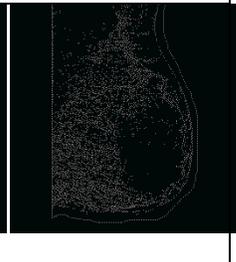 |
| Prewitt | 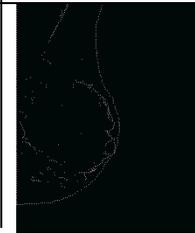 | 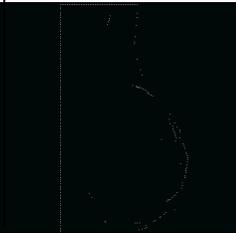 | 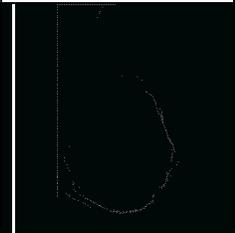 | 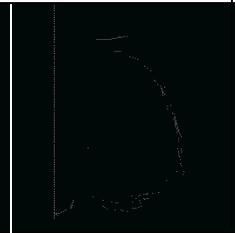 | 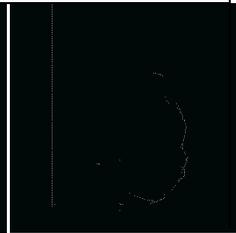 |

Fig. 3 Different Edge Detection Methods and Proposed Fuzzy based Edge Detection Methods using mdb002, mdb067, mdb171, mdb240 and mdb320 mammogram images